\pdfoutput=1

\documentclass[11pt]{article}

\usepackage{ACL2023}

\usepackage{times}
\usepackage{latexsym}
\usepackage{multicol}
\usepackage{multirow}
\usepackage{colortbl}
\usepackage{array}
\usepackage{tabularx}

\usepackage{soul}
\usepackage{arydshln}
\usepackage{booktabs}

\usepackage[utf8]{inputenc}
\usepackage[T1]{fontenc}
\usepackage{graphicx}
\usepackage{enumitem}

\usepackage{microtype}
\usepackage{inconsolata}
\usepackage{soul}

\newcolumntype{P}[1]{>{\centering\arraybackslash}p{#1}}
\ifx \final \undefined
\newcommand{\hs}[1]{\textcolor{blue}{Huan: #1}}
\newcommand{\xd}[1]{\textcolor{orange}{Xiang: #1}}

\newcommand{\ww}[1]{\textcolor{green}{Will: #1}}

\newcommand{\bs}[1]{\textcolor{red}{Boshi: #1}}
\newcommand{\nop}[1]{}
\else
\newcommand{\hs}[1]{}
\newcommand{\xd}[1]{}
\newcommand{\ww}[1]{}

\newcommand{\bs}[1]{}
\newcommand{\nop}[1]{}
\fi

\newcommand{\bobj}[1]{\textcolor{cyan}{#1}}
\newcommand{\ltemp}[1]{\textcolor{magenta}{#1}}
\newcommand{\cohere}[1]{\textit{#1}}
\newcommand{\rel}[1]{\underline{#1}}
\newcommand\ttsmall[1]{\texttt{\textmd {#1}}}
\usepackage{tikz}
\newcommand*\circled[1]{\tikz[baseline=(char.base)]{
            \node[shape=circle,draw,inner sep=1pt] (char) {\small{#1}};}}

\title{Towards Understanding Chain-of-Thought Prompting:\\  An Empirical Study of What Matters}

\newcommand{\aff}[1]{\rlap{\textsuperscript{\normalfont#1}}}
\author{
    Boshi Wang\aff{1} \hspace{1.2em}
    Sewon Min\aff{2} \hspace{1.2em}
    Xiang Deng\aff{1} \hspace{1.2em}
    Jiaming Shen\aff{3} \hspace{1.2em}
    You Wu\aff{3} \\
    {\bf Luke Zettlemoyer}\aff{2} \qquad
    {\bf Huan Sun}\aff{1} \\
    $^1$The Ohio State University \qquad
    $^2$University of Washington \qquad 
    $^3$Google Research \\
    \texttt{\{wang.13930,deng.595,sun.397\}@osu.edu} \\
    \texttt{\{sewon,lsz\}@cs.washington.edu}, \texttt{\{jmshen,wuyou\}@google.com}
}

\begin{document}
\maketitle
\begin{abstract}
Chain-of-Thought (CoT) prompting can dramatically improve the multi-step reasoning abilities of large language models (LLMs). CoT explicitly encourages the LLM to generate intermediate rationales for solving a problem, by providing a series of reasoning steps in the demonstrations. Despite its success, there is still little understanding of what makes CoT prompting effective and which aspects of the demonstrated reasoning steps contribute to its performance. In this paper, we show that CoT reasoning is possible even with invalid demonstrations---prompting with invalid reasoning steps can achieve over 80-90\% of the performance obtained using CoT under various metrics, while still generating coherent lines of reasoning during inference. Further experiments show that other aspects of the rationales, such as being relevant to the query and correctly ordering the reasoning steps, are much more important for effective CoT reasoning. Overall, these findings both deepen our understanding of CoT prompting, and open up new questions regarding LLMs' capability to learn to reason in context.\footnote{Our code and model input/output are available \href{https://github.com/sunlab-osu/Understanding-CoT}{here}.}
\end{abstract}

\section{Introduction}

\begin{figure}[ht]
  \centering
    \includegraphics[width=0.99\linewidth]{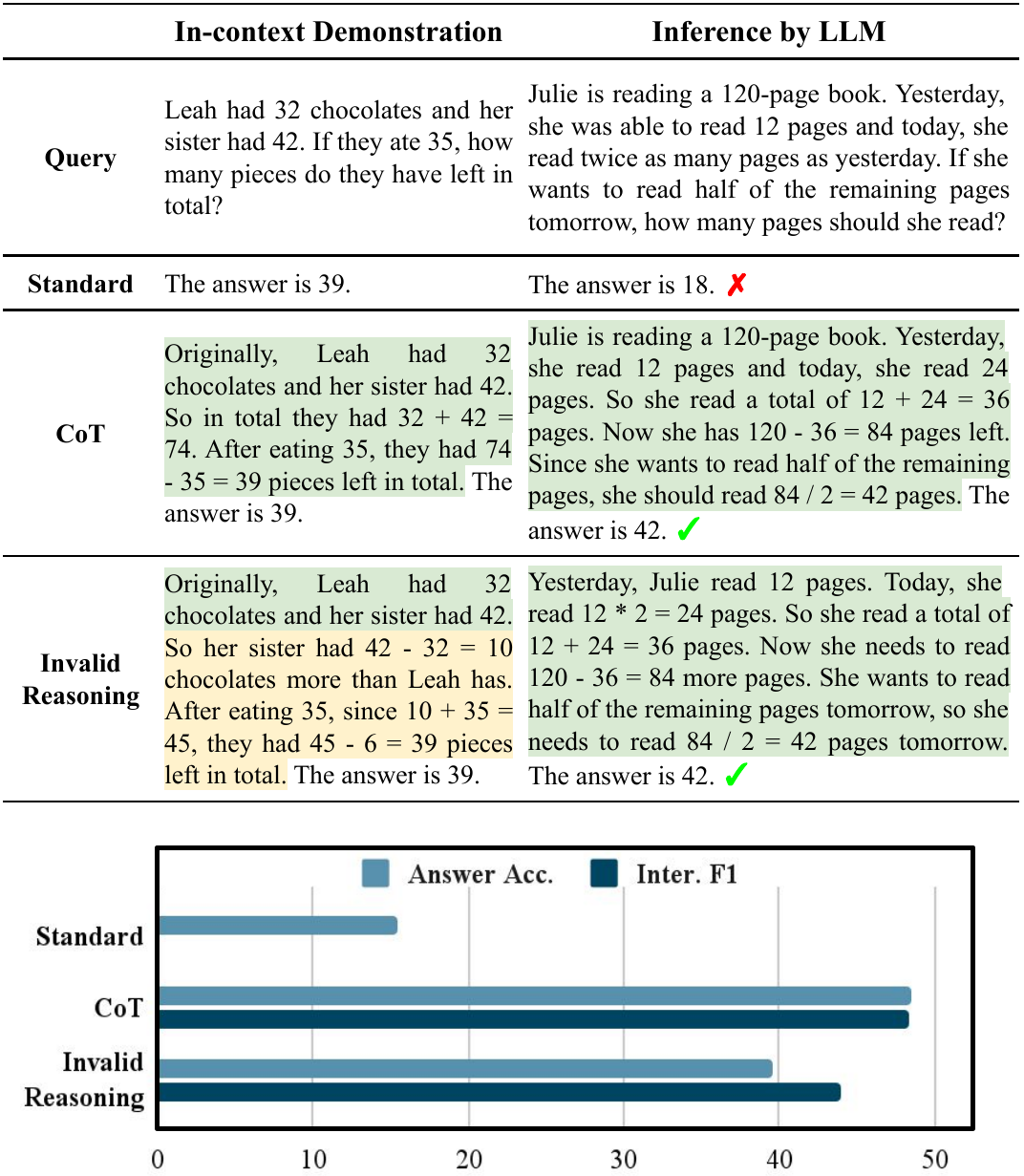}
  \caption{Results of standard prompting, Chain-of-Thought (CoT) prompting, and our ablation setting with invalid reasoning (\S\ref{sec:invalid reasoning}). We show one demonstration example and one inference example for arithmetic reasoning, where the rationale is in color (green: valid, yellow: invalid). We find that valid reasoning for the demonstrations matters only a small portion to the performance of CoT---by providing rationales with invalid reasoning, LLMs can achieve over 80-90\% of the performance of CoT under various metrics while performing logically sound and pertinent reasoning.}
\label{fig:CoT}
\end{figure}

Large language models (LLMs) can perform new tasks during inference when prompted with a few demonstrations~\cite{brown2020language}. Chain-of-Thought (CoT) prompting \cite{wei2022chain} can (Figure \ref{fig:CoT}) improve the ability of sufficiently large LLMs to do complex and multi-step reasoning. In addition to (query, answer) example-pair demonstrations, CoT prompting  includes a \textit{rationale} (colored part in Figure \ref{fig:CoT}) for each example, i.e., a series of reasoning steps towards the answer, which encourages the LLM to explicitly generate its intermediate reasoning process before predicting the final answer. Despite its successes, there is little understanding of what makes CoT prompting effective and which aspects of the demonstrated reasoning steps contribute to its performance. Recent findings also reveal that in-context learning could be very different from fine-tuning/training; for example, \citet{min2022rethinking} and \citet{webson-pavlick-2022-prompt} show that providing random labels or misleading instructions in context only marginally harms model performance for certain tasks. Inspired by this work, we take a closer look at CoT prompting to study how and why it works.

We design a series of ablation experiments where we deliberately change different aspects of the demonstrated rationales and measure how the model performance varies accordingly (\S\ref{sec:invalid reasoning}, \S\ref{sec:matters}). On two representative multi-step reasoning tasks---arithmetic reasoning and multi-hop factual question answering (QA), we find that \textbf{the validity of reasoning matters only a small portion to the performance}---by providing rationales with completely invalid reasoning steps, the LLM can still achieve over 80-90\% of the performance of CoT under various metrics while generating coherent lines of reasoning towards the answer (\S\ref{sec:invalid reasoning}). Through further examinations, we identify and formulate other aspects of a CoT rationale (\S\ref{sec:matters}), and find that \textbf{being relevant to the query and correctly ordering the reasoning steps are the key} for the effectiveness of CoT prompting. 

Overall, our findings suggest that what LLMs \textit{learn} about how to reason under CoT prompting could be limited. Rather, they have already gained a lot of such ``reasoning abilities'' from pretraining, and the demonstrations may mainly specify an output space/format that regularizes the model generation to look step-by-step while being in order and relevant to the query. Our work suggests a new way of interpreting the evaluation scores in view of the prior knowledge LLMs possess, and leads to reflections on benchmarking few-shot reasoning which we discuss in \S\ref{sec:discussion}.

\section{Background \& Study Formulation}
\label{sec:components}

\noindent\textbf{Chain-of-Thought (CoT) prompting.} Different from the standard way of prompting language models where a set of (query, answer) pairs are given as demonstrations~\citep{brown2020language}, CoT prompting~\citep{wei2022chain} additionally includes a rationale (Figure \ref{fig:CoT}, colored) for each example, encouraging the model to verbalize the intermediate reasoning steps for solving the task. Such a technique has been shown to improve the performance of LLMs with sufficient scale on complex reasoning, sometimes to a large degree especially on arithmetic reasoning, multi-hop question answering, and symbolic reasoning.

\noindent\textbf{Components of a CoT rationale.} We identify two distinct components of a CoT rationale (examples in Table \ref{tbl:components}):
\begin{itemize}[leftmargin=*]
    \item \bobj{Bridging objects}:
    the key and necessary objects that the model needs to traverse in order to make a successful final prediction. For arithmetic reasoning, the bridging objects are defined to be the numeric part (numbers \& equations) of the rationale, and for factual QA, the bridging objects are defined to be the subject \& object entities. 
    \item \ltemp{Language templates}:
    the complementary parts of bridging objects, which serve as textual hints and relations/predicates that guide the model to derive the correct bridging objects along the reasoning process.  
\end{itemize}

\noindent\textbf{Research questions.} In Chain-of-Thought prompting, correct bridging objects and language templates are provided as demonstrations to show the LLM how to reason. While CoT achieves impressive performance, we are interested in the following questions: \textit{are ground truth bridging objects/language templates important? If not, what would be the key aspects that are needed for the LLM to reason properly?} These questions are the main focus of our study, which will be discussed in \S\ref{sec:invalid reasoning} and \S\ref{sec:matters}.

\begin{table}[t]
\footnotesize
\begin{tabularx}{0.48\textwidth}{XX}
    
     \toprule
     \multicolumn{1}{c}{\bf Arithmetic Reasoning} & \multicolumn{1}{c}{\bf Multi-hop QA} \\
     \midrule
     Q: Leah had 32 chocolates and her sister had 42. If they ate 35, how many pieces do they have left in total?  & Q: Who is the grandchild of Dambar Shah?  \\
     \midrule
      A: \ltemp{Originally, Leah had \bobj{32} chocolates and her sister had \bobj{42}. So in total they had \bobj{32 + 42 = 74}. After eating \bobj{35}, they had \bobj{74 - 35 = 39} pieces left in total.} The answer is \bobj{39}. &  A: \ltemp{\bobj{Dambar Shah (? - 1645)} was the father of \bobj{Krishna Shah}. \bobj{Rudra Shah} was the child of \bobj{Krishna Shah (? - 1661)}.} So the final answer \ltemp{(the name of the grandchild)} is: \bobj{Rudra Shah}.  \\ 
     \bottomrule
\end{tabularx}
\caption{\bobj{Bridging objects} and \ltemp{language templates} of a Chain-of-Thought rationale. Here we illustrate with one in-context exemplar for each task we experiment with.}
\label{tbl:components}
\end{table}

\section{Experimental Setup}
\label{sec:exp setup}
\subsection{Datasets \& In-context Exemplars}
We experiment on two representative tasks involving multi-step reasoning: arithmetic reasoning \& multi-hop factual question answering (QA). We select benchmarks on which CoT prompting brings significant improvements over standard prompting, as shown in previous work \cite{wei2022chain, press2022measuring}; they are more suitable for our study, since our goal is to understand how different aspects of the Chain-of-Thought rationales contribute to the performance of CoT prompting. For arithmetic reasoning, we experiment on GSM8K \cite{cobbe2021training}, one of the most challenging mathematical reasoning benchmarks available which is also repeatedly adopted by prior work as a key benchmark for arithmetic reasoning; for multi-hop factual QA, we experiment on Bamboogle, a dataset of compositional questions constructed by \citet{press2022measuring}. Due to budget considerations, we uniformly sample 800 out of the 1319 test examples for GSM8K for evaluation. We evaluate on all 125 test samples for Bamboogle.

We base our experiments on the original prompt exemplars, i.e., the set of (query, rationale, answer) pairs released by \citet{wei2022chain} and \citet{press2022measuring}, with slight editing to make the structure more consistent and reduce redundancy, which makes our ablations more convenient to conduct. These edits only slightly affect the performance of CoT; we show our edited demonstration examples and include more details in Appendix \ref{app:CoT exemplars}.

\subsection{Backbone Language Model}
We use InstructGPT-175B\footnote{We also tried the original GPT-3 175B without instruction-finetuning in our preliminary experiments, but find that CoT prompting does not yield much performance gain than standard prompting, echoing \citet{Fu2022how}.}~\cite{ouyang2022training,brown2020language} \ttsmall{text-davinci-002} as our backbone LLM, which
is one of the most performant and widely-used LLMs with public APIs and has demonstrated strong performance under CoT prompting \cite{wei2022chain}. We report its results and analyze them
in the main content. In addition, we also test on \ttsmall{text-davinci-003} (a very recent improved version of \ttsmall{text-davinci-002}), PaLM~\citep{chowdhery2022palm} and Flan-PaLM~\citep{chung2022scaling}, where the results and discussion could be found in Appendix \ref{app:result-003}. All generations are done by greedy decoding (i.e., sampling with zero temperature) as in the original CoT work \cite{wei2022chain}.

\subsection{Evaluation}
\label{sec:evaluation}
Prior work mainly performs evaluation using the correctness of the final answer, which could be viewed as an \textit{extrinsic} way of assessing the predicted rationales. However, this may not align well with the actual quality of the rationale in many cases, as mentioned in \citet{huang2022towards}. For example, a rationale that is correct for all but the last step (and hence derives the wrong final answer) would still be assigned a zero score, while a rationale that is wrong/incomplete but reaches the correct final answer would be assigned a full score. Therefore, in addition to extrinsic evaluation (\textbf{Answer Accuracy} for GSM8K, \textbf{Answer F1} for Bamboogle), we perform \textit{intrinsic} evaluation where we measure the Recall/F1 (\textbf{Inter.\footnote{Abbreviation for ``Intermediate''.} Recall/F1}) of the bridging objects which need to be derived by the LLM (i.e., those that do not appear in the query). 
For GSM8K, since annotations for ground truth reasoning steps are available, we use the derived numbers in the annotated steps as a proxy for bridging objects.\footnote{We do not use whole equations since we observe that the LLM may express the mathematical equation in different ways, e.g., ``5 plus 3 is 8'', ``5 + 3 = 8''.} For Bamboogle, we manually annotate the bridging objects (intermediate entities) and measure their recall. While it is still possible for the model to reach correct bridging objects with the wrong language templates, we manually verify that this rarely happens; details are included in Appendix \ref{app:eval}. 


\section{How Much Does Valid Reasoning Matter?}
\label{sec:invalid reasoning}



Intuitively, one of the most important aspects of a Chain-of-Thought rationale would be its logically valid and sound reasoning. If we provide rationales with invalid reasoning steps in the demonstrated examples instead, we should expect the LLM to fail to reason properly and gain little or even negative improvements compared with standard prompting (where no rationale is given), since we are teaching the LLM to reason in the wrong way which could be even worse than not doing so at all. To test this intuition, we design an ablation study where we construct invalid reasoning steps for the demonstrated rationales, and measure its influence on model behavior.

\subsection{Constructing Invalid Chain of Reasoning}



We manually write rationales with invalid reasoning for \textit{all} the in-context demonstration examples. Since our focus here is to investigate the importance of the validity of reasoning, we only ablate the parts in a CoT rationale\nop{the CoT rationales} which are involved with derivations that are logically sound and helpful for answering the query. More specifically, we keep the premise steps which are copies/paraphrases of facts from the query, and change the subsequent steps such that they do not logically derive the final answer. Importantly, we are \textit{not} adopting an adversarial/counterfactual perturbation setting where minimal alterations are applied to make the reasoning invalid; instead, we apply rather drastic changes where we change both the bridging objects and language templates and hence little valid reasoning exists to help solve the query. The full prompts in our setting are included in Appendix \ref{app:full prompts}.

For example, consider an in-context demonstration (see \circled{1} in Table \ref{tbl:examples}) for arithmetic reasoning. Here the query is \textit{``Leah had 32 chocolates and her sister had 42. If they ate 35, how many pieces do they have left in total?''}. For the 1st entailment step which should sum \textit{``32''} and \textit{``42''} to get the total amount \textit{``32 + 42 = 74''} as in CoT, we instead write \textit{``So her sister had 42 - 32 = 10 chocolates more than Leah has.''} which has both the wrong bridging object and language template, and is completely unhelpful for solving the problem. The subsequent steps are written based on the previous steps, and in the end, answer the question whereas the rationale does not in any way lead to the answer logically. While the step itself still describes something that could be entailed in the example we just gave, this is not the case generally and most of the steps we write are neither helpful nor entailments from earlier steps. For example, the next step \textit{``After eating 35, since 10 + 35 = 45, they had 45 - 6 = 39 pieces left in total''} makes use of unwarranted information (\textit{``6''}) and has no valid entailment anywhere. We illustrate our construction using another example for factual QA, where the question is \textit{``Who is the grandchild of Dambar Shah?''}. Here, we write a rationale that finds the kingdom of \textit{``Dambar Shah''} and then a child of the person who established the kingdom, which does not lead to \textit{``the grandchild of Dambar Shah''}.

\begin{table*}[htbp]
\centering
\resizebox{\linewidth}{!}{
\begin{tabular}{lccccc}
\toprule
     &\multicolumn{3}{c}{\textbf{GSM8K}}&\multicolumn{2}{c}{\textbf{Bamboogle}}  \\
     \cmidrule(l{0.5em}r{0.5em}){2-4}\cmidrule(l{0.5em}r{0.5em}){5-6}
     &\textbf{Inter. Recall}&\textbf{Inter. F1}&\textbf{Answer Acc.}&\textbf{Inter. Recall}&\textbf{Answer F1}\\
     \midrule
     STD (Standard prompting)& N/A & N/A& 15.4 & N/A & 20.6\\
     CoT (Chain-of-Thought prompting) &43.9 & 48.3 & 48.5 & 45.2& 45.2 \\
     \midrule
     \circled{1} Invalid Reasoning &39.8 & 43.9 & 39.5 & 44.4 & 39.4 \\
     \midrule
     \circled{2} No \cohere{coherence} for \bobj{bridging objects} & 35.3 & 39.2 & 35.8 & 40.8 & 37.4 \\
     \circled{3} No \rel{relevance} for \bobj{bridging objects} & 21.4 & 26.2 &27.5 & 39.6 & 34.0 \\
     \circled{4} No \cohere{coherence} for \ltemp{language templates} & 24.1 & 28.3 &25.8 & 35.2 & 32.1 \\
     \circled{5} No \rel{relevance} for \ltemp{language templates} & 29.5 & 34.0 &32.8 & 40.4 & 29.4\\
     \circled{6} No \cohere{coherence} & 25.2 & 29.4 &23.1 & 39.6 & 33.8 \\
     \circled{7} No \rel{relevance} & 9.6 & 11.9 &11.0 & 36.8 & 23.9 \\
     \bottomrule
\end{tabular}}
\caption{Intrinsic and extrinsic evaluation results under InstructGPT (\ttsmall{text-davinci-002}) for all settings in our experiments. Results for \ttsmall{text-davinci-003}, PaLM and Flan-PaLM could be found in Appendix \ref{app:result-003}.}
\label{tbl:results}
\end{table*}

\subsection{Results \& Analysis}

\noindent\textbf{Quantitative results.} Table \ref{tbl:results} summarizes the quantitative results for \ttsmall{text-davinci-002}. We include additional results and discussion for \ttsmall{text-davinci-003}, PaLM and Flan-PaLM in Appendix \ref{app:result-003}. LLMs can achieve surprisingly high performance when provided with invalid reasoning steps for the demonstrations (\circled{1}). In particular, under \textbf{Inter. Recall}/\textbf{Inter.F1}, i.e., intrinsic evaluation, which is arguably a more faithful measurement of the rationale quality (\S\ref{sec:evaluation}), all LLMs we tested can retain over 90\% of the performance achieved under CoT prompting.

\begin{figure}[h]
  \centering
    \includegraphics[width=0.98\linewidth]{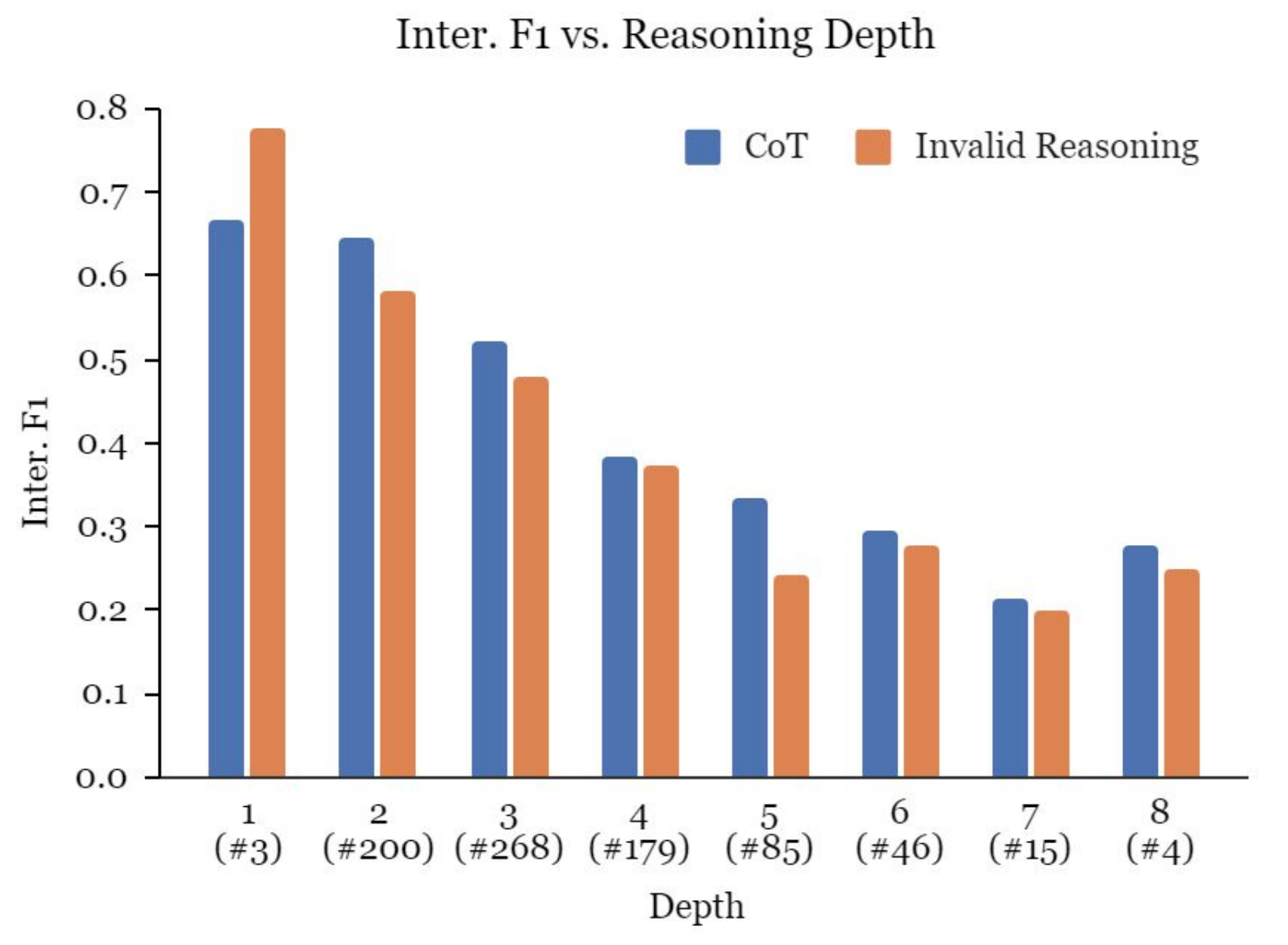}
  \caption{Model performance using CoT and demonstrations with invalid reasoning for examples with different reasoning depths on GSM8K. The number of samples for each reasoning depth is shown below (led by \textit{``\#''}). The performance drop is consistent across different levels of difficulty.}
\label{fig:depth_split}
\end{figure}

For GSM8K where there are large variations in the difficulty levels (here, we use the number of reasoning steps required to solve a problem as its difficulty level) of the problem instances, we additionally examine the model performance separately for each difficulty level. The results are shown in Figure \ref{fig:depth_split}. The performance drop is also uniform across samples with different levels of difficulty. At the instance level, after omitting samples where both settings get the correct/wrong answer, there is a significant portion for the remaining ones (62/196 for GSM8K, 6/20 for Bamboogle) where CoT gets the wrong answer and the invalid reasoning setting gets the correct answer. This further strengthens the finding that there is no strong connection between the reasoning validity of the demonstrations and the quality of the model predictions. 

\begin{table}[htbp]
\centering
\resizebox{\linewidth}{!}{
\begin{tabular}{lP{2cm}P{2cm}}
\toprule
     \textbf{Error Types} & \textbf{CoT} correct \& \textbf{IR} wrong &  \textbf{CoT} wrong \& \textbf{IR} correct\\
     \midrule
     Calculation & 20\% & 20\%\\
     \midrule
     One step missing & 35\% & 25\% \\
     \midrule
     Semantic understanding & 45\% & 55\% \\
     \bottomrule
\end{tabular}}
\caption{Distribution of error types of 20 examples from GSM8K where Chain-of-Thought (\textbf{CoT}) prompting reaches the correct answer and the Invalid Reasoning setting (\textbf{IR}) reaches a wrong answer, and 20 examples for the opposite case. }
\label{tbl:qual}
\end{table}

\noindent\textbf{Qualitative analysis.} By checking the generated rationales for the invalid reasoning setting, we find that overall they look indistinguishable from the rationales generated by CoT prompting. In almost all cases where the predicted final answer is correct, the rationales do reach the answer with valid and sound reasoning steps (as in CoT), drastically different from those in the given demonstrations; for cases where the final answer is wrong, the errors the LLM makes are also in the same types with the errors made under CoT prompting. To compare the distribution of errors between CoT and the invalid reasoning setting, we examine 20 samples from GSM8K where CoT gets the correct final answer and the invalid reasoning setting gets the wrong answer, and another 20 examples for the opposite case. 
We use the same error categorizations as in \citet{wei2022chain} for the qualitative analysis, and summarize the results in Table \ref{tbl:qual}. 
The distributions of errors in both cases are highly similar. 

\noindent\textbf{Summary.} Combining the quantitative and qualitative results, we can see that there is a low chance for any systematic difference between CoT and the invalid reasoning setting to exist. The LLM still tries and manages to generate logically sound and pertinent reasoning decently, and ablating the validity of reasoning for the demonstrations only brings a small performance degradation. 
This opens the question: \textit{If valid reasoning is not required, what are the key aspects that determine the effectiveness of CoT prompting?}

\section{What are the Key Aspects of Chain-of-Thoughts?}
\label{sec:matters}
Re-examining the rationales in our ablation setting in \S\ref{sec:invalid reasoning}, we can find that even though the reasoning is invalid, they have the following properties:
\begin{itemize}[leftmargin=*]
    \item The rationales still use information from the query; more specifically, they still start from bridging objects mentioned in the query, and the language templates are related to the query. Recall our running example for arithmetic reasoning (Table \ref{tbl:examples}), even though the reasoning here is wrong, the numbers \textit{``32''} and \textit{``42''} are kept from the query, and the language templates are still about \textit{``Leah''}, \textit{``Leah's sister''} and \textit{``Chocolates''}, and try to seek the answer to the query. Therefore, the rationale is still relevant to the query being asked.
    \item Each step of a rationale still follows the previous steps. Using again the same example, the bridging object (equation in this case) \textit{``42 - 32 = 10''} in the first entailment step uses numbers from previous steps; likewise, the language template \textit{``So her sister had \_ chocolates more than Leah has''} is something that follows after the earlier steps. Hence, overall, the rationale still appears to be coherent.
\end{itemize}
We formulate two notions that capture these two aspects of a rationale in what follows. 

\noindent\textbf{Relevance.} A component of the rationale has relevance if it is based on the corresponding component from the query. For bridging objects, this could be formally defined as using the exact same objects mentioned in the query (numbers for arithmetic reasoning and entities for factual QA); for language templates, they have relevance if they are still about the same set of entities/relations as the query, and allude to the question being asked. For example, a template about \textit{``Patricia''} and \textit{``hair''} would not have relevance to a query about \textit{``Leah''} and \textit{``Chocolates''}, and similarly, a template that attempts to find the \textit{``brother-in-law''} of the topic entity does not have relevance to a query which seeks the \textit{``grandchild''} (Table \ref{tbl:examples}).

\noindent\textbf{Coherence.} A component of the rationale has coherence if it is in the correct order, i.e., later steps could not be pre-conditions for earlier steps and reversely, earlier steps could not be based on later steps. For example, a rationale where \textit{``32 + 42 = 74''} appears before the introduction of \textit{``32''} or \textit{``42''} would not have coherence on bridging objects, and similarly for language templates.

In what follows, we design a set of ablation settings to examine the impact of these two aspects for different components of a CoT-like rationale.

\begin{table*}[ht]
\centering
\footnotesize
\resizebox{\linewidth}{!}{
\begin{tabular}{p{3.6cm}p{6.7cm}p{6.7cm}}
     \toprule
     \textbf{Prompt Setting} & \textbf{Example Query (Arithmetic Reasoning)} \newline  \textit{Leah had 32 chocolates and her sister had 42. If they ate 35, how many pieces do they have left in total?}  & \textbf{Example Query (Factual QA)} \newline \textit{Who is the grandchild of Dambar Shah?}  \\
     \midrule
     STD (Standard prompting) &  39  &  So the final answer is: Rudra Shah.  \\
     \midrule
     \multirow{1}{*}{CoT (Chain-of-Thought)} &  Originally, Leah had 32 chocolates and her sister had 42. So in total they had 32 + 42 = 74. After eating 35, they had 74 - 35 = 39 pieces left in total. The answer is 39. &  Dambar Shah (? - 1645) was the father of Krishna Shah. Rudra Shah was the child of Krishna Shah (? - 1661). So the final answer (the name of the grandchild) is: Rudra Shah.  \\ 
     \midrule
     \circled{1} Invalid Reasoning &  Originally, Leah had 32 chocolates and her sister had 42. So her sister had 42 - 32 = 10 chocolates more than Leah has. After eating 35, since 10 + 35 = 45, they had 45 - 6 = 39 pieces left in total. The answer is 39.  &  Dambar Shah (? - 1645) was the king of the Gorkha Kingdom. The Gorkha Kingdom was established by Prince Dravya Shah. Dravya Shah has a child named Rudra Shah. So the final answer (the name of the grandchild) is: Rudra Shah. \\
     \midrule 
     \circled{2} No \cohere{coherence} for \bobj{bridging objects} &  Originally, Leah had 32 + 42 = 74 chocolates and her sister had 32. So in total they had 74 - 35 = 39. After eating 35, they had 42 pieces left in total. The answer is 39.  &  Krishna Shah was the father of Rudra Shah. Dambar Shah (? - 1645) was the child of Krishna Shah (? - 1661). So the final answer (the name of the grandchild) is: Rudra Shah.  \\
     \midrule
     \circled{3} No \rel{relevance} for \bobj{bridging objects} &  Originally, Leah had 19 chocolates and her sister had 31. So in total they had 19 + 31 = 50. After eating 29, they had 50 - 29 = 21 pieces left in total. The answer is 21. &  Metis Amando was the father of David Amando. Randall Amando was the child of David Amando. So the final answer (the name of the grandchild) is: Randall Amando. \\
     \midrule
     \circled{4} No \cohere{coherence} for \ltemp{language templates} &  After eating 32, they had 42 pieces left in total. Originally, Leah had 32 + 42 = 74 chocolates and her sister had 35. So in total they had 74 - 35 = 39. The answer is 39.  &  Dambar Shah (? - 1645) was the child of Krishna Shah. Krishna Shah (? - 1661)  was the father of Rudra Shah. So the final answer (the name of the grandchild) is: Rudra Shah. \\  
     \midrule
     \circled{5} No \rel{relevance} for \ltemp{language templates} &  Patricia needs to donate 32 inches, and wants her hair to be 42 inches long after the donation. Her hair is 35 inches long currently. Her hair needs to be 32 + 42 = 74 inches long when she cuts it. So she needs to grow 74 - 35 = 39 more inches. The answer is 39. &  The husband of Dambar Shah (? - 1645) is Krishna Shah. Krishna Shah (? - 1661) has a brother called Rudra Shah. So the final answer (the name of the brother-in-law) is: Rudra Shah. \\  
     \midrule
     \circled{6} No \cohere{coherence} &  After eating 32 + 42 = 74, they had 32 pieces left in total. Originally, Leah had 74 - 35 = 39 chocolates and her sister had 35. So in total they had 42. The answer is 39.  &  Krishna Shah was the child of Rudra Shah. Dambar Shah (? - 1645) was the father of Krishna Shah (? - 1661). So the final answer (the name of the grandchild) is: Rudra Shah. \\ 
     \midrule
     \circled{7} No \rel{relevance} &   Patricia needs to donate 19 inches, and wants her hair to be 31 inches long after the donation. Her hair is 29 inches long currently. Her hair needs to be 19 + 31 = 50 inc long when she cuts it. So she needs to grow 50 - 29 = 21 more inches. The answer is 21.  &  The husband of Metis Amando is David Amando. David Amando has a brother called Randall Amando. So the final answer (the name of the brother-in-law) is: Randall Amando. \\  
     \bottomrule
\end{tabular}
}
\caption{Examples for all settings in our experiments.}
\label{tbl:examples}
\end{table*}

\subsection{Ablation Settings}
In order not to introduce mixed effects which could make the results not well-controlled, we base the ablation settings on top of the CoT prompts instead of the setting in \S\ref{sec:invalid reasoning}. 

Given the two components (bridging objects and language templates) and the two aspects (relevance and coherence) of the rationale, there are naturally four ablation settings where each could examine one aspect of a certain component. We also experiment with two other settings: \rel{no relevance} where neither bridging objects nor language templates have relevance, and \cohere{no coherence} which is defined analogously (\circled{6}, \circled{7} in Table \ref{tbl:examples}). 

\noindent\textbf{Destroying relevance.} We perform random substitutions to ablate the relevance of a certain component. For ablating the relevance of bridging objects, we randomly sample alternatives (numbers for GSM8K, entities for Bamboogle) for those from the query, and change the bridging objects in the subsequent steps correspondingly to maintain the coherence of the rationale. Using our running example, we randomly replace the bridging objects from the query: \textit{``32''}$\rightarrow$\textit{``19''}, \textit{``42''}$\rightarrow$\textit{``31''} and \textit{``35''}$\rightarrow$\textit{``29''}, then change the bridging object from the first entailment step from \textit{``32 + 42 = 74''} to \textit{``19 + 31 = 50''}, and so on so forth. To ablate the relevance of language templates, for GSM8K, we randomly sample an annotated rationale from the training set, and use its template in place of the original template. For Bamboogle, we manually replace the template with an alternative which is irrelevant to the query.

\noindent\textbf{Destroying coherence.} Ablating the coherence is rather straightforward, where we randomly shuffle the components and permute their orderings.

\subsection{Results \& Analysis}
The results could be found in Table \ref{tbl:results}, and we include additional results for \ttsmall{text-davinci-003}, PaLM and Flan-PaLM in Appendix \ref{app:result-003}. We summarize the main findings in what follows. 

\noindent\textbf{Relevance and coherence are key for the performance of CoT prompting.} 
It can be seen that most of the settings for this section (\circled{2}-\circled{7}) have rather large performance drops from CoT, where the low-performing ones approach or even underperform standard prompting. This suggests that overall, relevance and coherence are key for the performance of CoT.


\noindent\textbf{Keeping relevance is crucial.} The no relevance setting \circled{7} where both components of the rationale have no relevance achieves significantly poorer performance than other ablation settings, and even underperforms standard prompting (STD) where no rationale is given on GSM8K. To see why such low performance happens, we manually examine the generated rationales under this setting for 20 examples on GSM8K. We find that the LLM is indeed generating irrelevant rationales (both bridging objects and language templates) for 15 out of 20 examples. Many of the rationales have recurring topics (e.g., ``cats and dogs'', ``passengers and buses'') which we hypothesize are frequent patterns in the portion relevant to mathematics in the pretraining corpora. Overall, this suggests that a certain level of relevance is crucial for the LLM to stick to the query being asked.

\noindent\textbf{Relevance matters more than coherence for bridging objects.} Providing incoherent bridging objects (\circled{2}) achieves better performance than providing irrelevant bridging objects (\circled{3}), especially on the more challenging GSM8K dataset (39.2 \textit{v.s.} 26.2 \textbf{ Inter. F1}). which indicates that it is important for the bridging objects to be relevant, but not as important to have them in the right order to guide the LLM along the reasoning process. We quantitatively measure the coverage of bridging objects from the query for the generated rationales, and find that the settings with no relevance for bridging objects (\circled{3}, \circled{7}) do have significantly lower coverage (below 60\%) than other settings (around 80\%).

\noindent\textbf{Coherence of language templates is important.} Different from the coherence of bridging objects \circled{2}, the coherence of language templates \circled{4} matters a lot to the performance of CoT prompting. By examining the predicted rationales, we find that the LLM is indeed generating rationales with incoherent language templates (14 out of 20 examples), which negatively affects reasoning. 


\section{Discussion}
\label{sec:discussion}

The results from \S\ref{sec:invalid reasoning} and \S\ref{sec:matters} open up new questions regarding learning to reason in context for LLMs, which we discuss next.

\noindent\textbf{\textit{Do} LLMs learn to reason from CoT demonstrations?} Given the surprisingly high performance obtained by ablating the validity of reasoning for the in-context rationales, it can be concluded that what the LLM learns from the demonstrations about how to reason properly is limited---rather, the LLM has already gained a lot of such complex reasoning ability from pretraining (at least for tasks we experiment on), and the provided reasoning steps serve more as the role of an output format/space, that regularizes the LLM to generate rationales that look step-by-step while being coherent and relevant to the query. Moreover, results obtained from recent stronger models including \ttsmall{text-davinci-003} and Flan-PaLM (see Appendix \ref{app:result-003}) suggest that LLMs suffer further less from the ablations when they have more prior knowledge about the task. In particular, for Flan-PaLM which is directly trained on both arithmetic reasoning and factual QA in CoT fashion and hence has immense knowledge on these tasks~\citep{chung2022scaling}, it could be seen that none of the ablations has significant impacts on its performance. On the positive side, this indicates that LLMs can effectively utilize their prior knowledge to solve new problems. However, from another perspective, if we view the invalid reasoning setting as a \textit{task} where the goal is to generate invalid reasoning steps for the query, then the LLM has basically failed to capture the task as it still tries to predict valid reasoning steps. This leads to the concern that LLMs may over-rely on their prior knowledge and ignore important information in the context that are presumably rare in the pretraining distribution, including those that are crucial for specifying the task semantics \cite{jang2022can}.

\noindent\textbf{\textit{Can} LLMs learn to reason in-context?} We note that what we find does not in any way diminish the \textit{potential} of learning to reason in context for LLMs; recent work has also shown evidence that learning in context is possible and could be powerful \cite{garg2022can, akyurek2022learning}. Rather, our findings show that the existing successes of CoT are not sufficient for establishing that LLMs are good \textit{few-shot learners} of reasoning; instead, the pretraining corpora have already forged them to be good reasoners on the tasks being evaluated, and the main role that the demonstrations play is to elicit such reasoning skills.

\noindent\textbf{Reflections on benchmarking few-shot reasoning.} An important topic on benchmarking in the era of large pre-trained language models is to quantify the level of prior knowledge the LLM has gained about the end task being evaluated, which is crucial for assessing how well can the model truly extrapolate from pretraining and acquire new skills \cite{chollet2019measure}. One direct way is to look into the pretraining corpora when it is accessible, e.g., \citet{razeghi2022impact} investigates the correlation between the model performance and the frequency of terms from the test instances in the pretraining data. However, the pretraining corpora are not always accessible, and low-level statistics are usually not adequate when the topics of interest are abstract and high-level skills such as reasoning. Along this direction, our work could be regarded as a way to approximately quantify the prior knowledge that the LLM possesses on multi-step reasoning. Our findings indicate that evaluations on alternative benchmarks where LLMs have less prior knowledge are needed to more faithfully assess the LLMs' abilities on learning to reason from few-shot demonstrations.



\section{Related Work}
There have been several subsequent work of Chain-of-Thought prompting since its introduction. 
\citet{wang2022self} proposes to sample a diverse set of reasoning paths instead of performing greedy decoding, and marginalize over the sampled paths to select the most consistent answer. \citet{zhang2022automatic} proposes a method for automatically constructing the in-context exemplars for CoT. \citet{chen2022program} explores program-based CoT which can better disentangle computation from reasoning. In this paper, we are primarily focused on understanding the effectiveness of the original CoT prompting method where we use the same experimental settings (e.g., greedy decoding) and base our experiments on the same few-shot exemplars used. We believe our findings could also apply to some of the subsequent variants of CoT prompting.

A few recent work focuses on understanding/analyzing CoT prompting. \citet{madaan2022text} investigates the importance of different components of the demonstrated CoT rationales by changing them to be \textit{counterfactual}. 
They only experiment with limited ways of changing the rationales to be \textit{wrong} including using incorrect calculations (e.g., \textit{``5 + 4 = 7''}) or entities. For most of their settings, even though the rationales are made counterfactual, they are still \textit{correct} since the query is changed accordingly (see, e.g., Table 48 of their paper). 
Concurrent to our work, \citet{ye2022complementary} also explores how the model performance could be affected by corrupting the CoT rationales. They experiment with using incorrect calculations and \textit{dropping} (parts of) the bridging objects/language templates, which are different from our ablation designs. \citet{saparov2022language} investigates systematically evaluating CoT by creating a synthetic QA dataset based on first-order logic, which allows for parsing the generated rationales into symbolic proofs for formal analysis. Overall, to our knowledge, we are the first to show that it is possible to have CoT rationales that are wrong and drastically deviate from the gold ones while still maintaining high model performance. 

In general in-context learning (ICL), \citet{min2022rethinking} shows that for a wide range of tasks in natural language understanding with categorical label space (classification and multi-choice), ground truth input-label mappings matter very little for end-task performance, and other aspects such as the label space, overall format and the distribution of text are the key. Building on this work, \citet{kim2022ground} finds that the correct input-label correspondence could have varying impacts based on the task and experimental configurations, and \citet{wei2023larger} finds that models with larger scale can override semantic priors and learn input-label mapping in context. \citet{webson-pavlick-2022-prompt} finds that for instruction models, the performance on natural language inference tasks has small degradations under irrelevant or misleading instructions. \citet{xie2022an} provides theoretical analysis of ICL by formulating it as Bayesian inference. Our work could be viewed as an attempt to empirically understand ICL in sequence generation tasks requiring multi-step reasoning.

\section{Conclusion}
\label{sec:conclusion}
In this paper, we aim to better understand Chain-of-Thought prompting through a series of ablation experiments that unveil the impact of different aspects of a CoT rationale. We find that 1) the validity of reasoning in the prompting examples matters only a small portion to the performance; 2) relevance to the input query and following the order along the reasoning steps are the key to the effectiveness of CoT prompting. Overall, our findings deepen the understanding of CoT prompting, and open up new questions/reflections regarding LLMs’ capability of learning to reason in context.

\section*{Limitations}
\noindent\textbf{Experiments on other types of reasoning tasks.} In addition to the two representative reasoning tasks (arithmetic reasoning and multi-hop question answering) that we experiment on, there are also other tasks where CoT prompting brings significant improvements over standard prompting shown by previous work, many of which are symbolic reasoning tasks such as {Last letter concatenation}, {Coin flip} from \citet{wei2022chain} and {Temporal Sequences}, {Tracking Shuffled Objects} from BIG-Bench~\cite{srivastava2022beyond,suzgun2022challenging}. However, most (if not all) tasks there are highly \textit{template-based} and hence the reasoning steps have little variations, both within each example and across different examples. This makes it difficult for us to conduct our ablation studies on these tasks. Take the example of {Last letter concatenation}, a task about concatenating the last letters of a given sequence of words (e.g., \textit{``Amy Brown''} $\rightarrow$ \textit{``yn''}). Here, every step in the rationale except the last is in the form \textit{``\ltemp{The last letter of} \bobj{X} \ltemp{is} \bobj{Y}''} where \textit{X} is some word in the given sequence and \textit{Y} is the last letter of \textit{X}. Hence, the language templates are the same and there is no sense of order among the steps (the order is completely characterized by the given sequence instead), and our ablation settings will not apply well. Extending our ablation designs to these ``reduced'' cases is one of the items we want to explore in the future.

\noindent\textbf{A more systematic treatment of ``invalid reasoning''.} We manually write rationales with invalid reasoning for the experiments in \S\ref{sec:invalid reasoning} since automatically synthesizing such rationales turns out to be challenging, mostly due to the informal nature of the tasks we experiment on (relatedly, the original CoT rationales are also human-written). We intend to give a more systematic treatment of the invalid reasoning setting in the future, e.g., following the categorizations of informal logical fallacies~\cite{copi2016introduction}.

\noindent\textbf{Improvements on intrinsic evaluation.} Our intrinsic evaluation of the generated rationales is based on the correctness of bridging objects, which, even though is a good indicator of the quality of language templates (Appendix \ref{app:eval}) in our experiments, may not be a good metric in general cases. It also relies on ground truth bridging objects, which are usually not available and costly to annotate. Toward this end, one direction we want to explore further is to develop ways to conduct more comprehensive and reference-free intrinsic evaluations. Recent papers such as \citet{golovneva2022roscoe} have also done promising work along this line. 

\section*{Acknowledgements}
The authors would like to thank the anonymous reviewers and colleagues from the OSU NLP group for their thoughtful comments. This research was supported in part by Google Faculty Award, Google Research Scholar Award, NSF IIS 1815674, NSF CAREER 1942980, NSF OAC-2112606, and Ohio Supercomputer Center~\citep{OhioSupercomputerCenter1987}. The views and conclusions contained herein are those of the authors and should not be interpreted as representing the official policies, either expressed or implied, of the U.S. government. The U.S. Government is authorized to reproduce and distribute reprints for Government purposes notwithstanding any copyright notice herein. 

\bibliography{anthology,custom}
\bibliographystyle{acl_natbib}

\onecolumn
\pagebreak
\twocolumn

\appendix
\section{Appendix}

\subsection{Chain of Thought Exemplars}
\label{app:CoT exemplars}
We base our experiments on the original prompt exemplars released by \citet{wei2022chain, press2022measuring} with slight editing to make the structure more consistent and reduce redundancy, which makes our ablations more convenient to conduct. The edited CoT prompts for arithmetic reasoning and multi-hop QA could be found in Table \ref{tbl:prompt cot arith} and Table \ref{tbl:prompt cot QA} respectively. We mainly perform the following edits: 1) shift premise steps (copy/paraphrase of facts from the query) to the beginning steps of the rationale; 2) add/expand the language templates for steps with no/over-concise language templates; 3) remove unnecessary steps/information that are unhelpful for answering the query. 

Overall, these edits only slightly affect the performance of CoT. A comparison of the performance is shown in Table \ref{app:tbl comp cot}. 

\begin{table*}[ht]
\centering
\resizebox{\linewidth}{!}{
\begin{tabular}{lccccc}
\toprule
     &\multicolumn{3}{c}{\textbf{GSM8K}}&\multicolumn{2}{c}{\textbf{Bamboogle}}  \\
     \cmidrule(l{0.5em}r{0.5em}){2-4}\cmidrule(l{0.5em}r{0.5em}){5-6}
     &\textbf{Inter. Recall}&\textbf{Inter. F1}&\textbf{Answer Acc.}&\textbf{Inter. Recall}&\textbf{Answer F1}\\
     \midrule
     Chain-of-Thought (Original)& 44.5 & 48.7 & 48.1 & 44.8 & 43.1\\
     Chain-of-Thought (After Editing) &43.9 & 48.3 & 48.5 & 45.2 & 45.2\\
     \bottomrule
\end{tabular}}
\caption{Performance comparison (under \ttsmall{text-davinci-002}) of the Chain-of-Thought exemplars before/after our editing.}
\label{app:tbl comp cot}
\end{table*}

\subsection{More Details on Intrinsic Evaluation}
\label{app:eval}
We use Recall/F1 of the bridging objects as the metrics for intrinsic evaluation of the generated rationales. While the metrics don't take into account the quality of the language templates, we examine the predicted rationales for 20 random examples under each setting we tested except standard prompting (which does not generate any rationale), and find that for all the examples, whenever the LLM reaches a correct bridging object, the corresponding language template within the step is also correct. This suggests that overall, the correctness of bridging objects is a very good indicator of the quality of the reasoning steps.

\subsection{Additional Results \& Discussion}
\label{app:result-003}

Table \ref{tbl:results-003} includes results for \ttsmall{text-davinci-003}, \ttsmall{text-davinci-002}'s very recent improved version. 

Comparing with the results from \ttsmall{text-davinci-002} (Table \ref{tbl:results}), it could be seen that \ttsmall{text-davinci-003} brings large performance improvements, especially under the ablation settings. In particular, providing invalid reasoning for the rationales (\circled{1}) overall only marginally harms the performance, and even outperforms CoT on GSM8K under intrinsic evaluation. This suggests that \ttsmall{text-davinci-003} is equipped with even stronger multi-step ``reasoning'' abilities on the evaluated tasks through pre-training, and learns little about how to reason from the demonstrations. 

For the remaining settings where we ablate the relevance/coherence (\circled{2}-\circled{7}), the same trend can be observed on the challenging GSM8K dataset, e.g., the model still suffers a lot when providing rationales that are irrelevant or have incoherent language templates. For the relatively easier Bamboogle dataset, the high model capacity indicated by its impressive performance has basically erased significant impacts from the ablations, with the only standing observation that the model still needs the rationales to be relevant to maintain its performance. 

Overall, from the performance achieved by \ttsmall{text-davinci-002} and \ttsmall{text-davinci-003}, we can observe a general trend where LLMs suffer less from the ablations when they have more prior knowledge about the task. To further explore this, we test on Flan-PaLM~\citep{chung2022scaling}, the instruction-tuned version of PaLM~\citep{chowdhery2022palm} that is directly trained on both arithmetic reasoning and factual QA in CoT fashion during instruction tuning, and hence has immense knowledge on these tasks. The results are shown in Table \ref{tbl:results-fpalm}. It could be seen that none of the ablations has significant impacts on the model performance, which further strengthens this pattern. On the positive side, this indicates that LLMs can effectively utilize their prior knowledge to solve new problems; however, this also leads to the concern that LLMs may over-rely on their prior knowledge and ignore important information in the context, including those that are crucial for specifying the task semantics \cite{jang2022can}.

We also test PaLM, which is a non-instruction-finetuned LLM that exhibits strong CoT reasoning ability. The results are included in Table \ref{tbl:results-palm}. Overall, similar observations could be found, which suggests that our findings are not exclusive to instruction-tuned models. There are some inconsistencies between the performance from PaLM and InstructGPT on Bamboogle, where the importance of coherence and relevance for bridging objects is flipped. This could be the consequence of instruction tuning, and differences in pretraining corpora and model scales. 

\begin{table*}[htbp]
\centering
\resizebox{\linewidth}{!}{
\begin{tabular}{lccccc}
\toprule
     &\multicolumn{3}{c}{\textbf{GSM8K}}&\multicolumn{2}{c}{\textbf{Bamboogle}}  \\
     \cmidrule(l{0.5em}r{0.5em}){2-4}\cmidrule(l{0.5em}r{0.5em}){5-6}
     &\textbf{Inter. Recall}&\textbf{Inter. F1}&\textbf{Answer Acc.}&\textbf{Inter. Recall}&\textbf{Answer F1}\\
     \midrule
     STD (Standard prompting)& N/A & N/A& 15.2 & N/A & 25.1\\
     CoT (Chain-of-Thought prompting) &48.4 & 53.1 &54.5 & 61.6 & 59.5\\
     \midrule
     \circled{1} Invalid Reasoning &50.2 & 53.5 & 51.5 & 60.8 & 56.4\\
     \midrule
     \circled{2} No \cohere{coherence} for \bobj{bridging objects} & 46.5 & 51.5 & 50.4& 59.2 & 55.2\\
     \circled{3} No \rel{relevance} for \bobj{bridging objects} & 32.5 & 38.3 & 47.2 & 60.4 &  56.9\\
     \circled{4} No \cohere{coherence} for \ltemp{language templates} & 37.8 &  43.3& 41.9 & 57.2 & 51.4\\
     \circled{5} No \rel{relevance} for \ltemp{language templates} & 44.6 & 49.9 &51.8& 62.4 & 59.3\\
     \circled{6} No \cohere{coherence} & 34.5 & 39.4& 31.0& 57.6 & 55.2\\
     \circled{7} No \rel{relevance} &  15.5 &  17.8& 16.2&  50.0 &  49.0\\
     \bottomrule
\end{tabular}}
\caption{Intrinsic and extrinsic evaluation results under \ttsmall{text-davinci-003} for all settings. Discussions are included in Appendix \ref{app:result-003}.}
\label{tbl:results-003}
\end{table*}
\begin{table*}[htbp]
\centering
\resizebox{\linewidth}{!}{
\begin{tabular}{lccccc}
\toprule
     &\multicolumn{3}{c}{\textbf{GSM8K}}&\multicolumn{2}{c}{\textbf{Bamboogle}}  \\
     \cmidrule(l{0.5em}r{0.5em}){2-4}\cmidrule(l{0.5em}r{0.5em}){5-6}
     &\textbf{Inter. Recall}&\textbf{Inter. F1}&\textbf{Answer Acc.}&\textbf{Inter. Recall}&\textbf{Answer F1}\\
     \midrule
     STD (Standard prompting)& N/A & N/A& 21.8 & N/A & 36.5\\
     CoT (Chain-of-Thought prompting) &72.2 & 73.0 &63.8 & 57.6 & 56.9\\
     \midrule
     \circled{1} Invalid Reasoning &71.8 & 72.6 & 64.4 & 55.6 & 52.8\\
     \midrule
     \circled{2} No \cohere{coherence} for \bobj{bridging objects} & 72.1 & 72.9 & 65.8& 51.6 & 49.3\\
     \circled{3} No \rel{relevance} for \bobj{bridging objects} & 71.1 & 71.9 & 64.6& 54.0 & 52.8\\
     \circled{4} No \cohere{coherence} for \ltemp{language templates} & 71.6 & 72.2 & 63.9& 54.0 & 52.0\\
     \circled{5} No \rel{relevance} for \ltemp{language templates}& 71.9 & 72.7 & 64.9& 55.2 & 53.5\\
     \circled{6} No \cohere{coherence} & 71.7 & 72.5 & 64.2& 54.4 & 54.0\\
     \circled{7} No \rel{relevance} & 70.7 & 71.6 & 64.5& 50.0 & 51.9\\
     \bottomrule
\end{tabular}}
\caption{Intrinsic and extrinsic evaluation results under Flan-PaLM~\cite{chung2022scaling}, the instruction-tuned version of PaLM for all settings. Discussions are included in Appendix \ref{app:result-003}.}
\label{tbl:results-fpalm}
\end{table*}
\begin{table*}[htbp]
\centering
\resizebox{\linewidth}{!}{
\begin{tabular}{lccccc}
\toprule
     &\multicolumn{3}{c}{\textbf{GSM8K}}&\multicolumn{2}{c}{\textbf{Bamboogle}}  \\
     \cmidrule(l{0.5em}r{0.5em}){2-4}\cmidrule(l{0.5em}r{0.5em}){5-6}
     &\textbf{Inter. Recall}&\textbf{Inter. F1}&\textbf{Answer Acc.}&\textbf{Inter. Recall}&\textbf{Answer F1}\\
     \midrule
     STD (Standard prompting)& N/A & N/A& 15.0 & N/A & 31.0\\
     CoT (Chain-of-Thought prompting) &36.6 & 40.6 &37.0 & 54.0 & 54.8\\
     \midrule
     \circled{1} Invalid Reasoning &33.9 &36.9 &31.8 & 50.4 & 46.1\\
     \midrule
     \circled{2} No \cohere{coherence} for \bobj{bridging objects} &30.3 & 35.0 &33.5 & 33.6 & 25.7\\
     \circled{3} No \rel{relevance} for \bobj{bridging objects} &15.5 & 20.1 &21.2 & 47.2 & 47.7\\
     \circled{4} No \cohere{coherence} for \ltemp{language templates} &23.1 & 27.3 &21.9 & 40.4 & 35.5\\
     \circled{5} No \rel{relevance} for \ltemp{language templates}&19.5 & 22.9 &20.4 & 38.4 & 30.6\\
     \circled{6} No \cohere{coherence} &23.9 & 28.3 &24.1 & 39.6 & 33.6\\
     \circled{7} No \rel{relevance} &12.1 & 16.4 &16.4 & 28.4 & 14.3\\
     \bottomrule
\end{tabular}}
\caption{Intrinsic and extrinsic evaluation results under PaLM. Discussions are included in Appendix \ref{app:result-003}.}
\label{tbl:results-palm}
\end{table*}

\subsection{Full List of Prompts}
\label{app:full prompts}
Full prompts for all settings in our experiments are included in Table \ref{tbl:prompt cot arith}-\ref{tbl:prompt NoRelev QA}.

\begin{table*}[ht]
\centering
\resizebox{0.95\linewidth}{!}{
\begin{tabular}{p{17cm}}
     \toprule
        \textbf{Q:} There are 15 trees in the grove. Grove workers will plant trees in the grove today. After they are done, there will be 21 trees. How many trees did the grove workers plant today? \\
        \textbf{A:} There are 15 trees originally. Then there were 21 trees after the Grove workers planted some more. So there must have been 21 - 15 = 6 trees that were planted. The answer is 6.\\
        \hdashline
        \textbf{Q:} If there are 3 cars in the parking lot and 2 more cars arrive, how many cars are in the parking lot?\\
        \textbf{A:} There are originally 3 cars. Then 2 more cars arrive. Now 3 + 2 = 5 cars are in the parking lot. The answer is 5.\\
        \hdashline
        \textbf{Q:} Leah had 32 chocolates and her sister had 42. If they ate 35, how many pieces do they have left in total?\\
        \textbf{A:} Originally, Leah had 32 chocolates and her sister had 42. So in total they had 32 + 42 = 74. After eating 35, they had 74 - 35 = 39 pieces left in total. The answer is 39.\\
        \hdashline
        \textbf{Q:} Jason had 20 lollipops. He gave Denny some lollipops. Now Jason has 12 lollipops. How many lollipops did Jason give to Denny?\\
        \textbf{A:} Jason had 20 lollipops originally. Then he had 12 after giving some to Denny. So he gave Denny 20 - 12 = 8 lollipops. The answer is 8.\\
        \hdashline
        \textbf{Q:} Shawn has five toys. For Christmas, he got two toys each from his mom and dad. How many toys does he have now?\\
        \textbf{A:} Shawn started with 5 toys. He then got 2 toys each from his mom and dad. So he got 2 * 2 = 4 more toys. Now he has 5 + 4 = 9 toys. The answer is 9.\\
        \hdashline
        \textbf{Q:} There were nine computers in the server room. Five more computers were installed each day, from monday to thursday. How many computers are now in the server room?\\
        \textbf{A:} There were originally 9 computers. For each day from monday to thursday, 5 more computers were installed. So 4 * 5 = 20 computers were added. Now 9 + 20 = 29 computers are now in the server room. The answer is 29.\\
        \hdashline
        \textbf{Q: }Michael had 58 golf balls. On tuesday, he lost 23 golf balls. On wednesday, he lost 2 more. How many golf balls did he have at the end of wednesday?\\
        \textbf{A:} Michael started with 58 golf balls. He lost 23 on Tuesday, and lost 2 more on wednesday. So he had 58 - 23 = 35 at the end of Tuesday, and 35 - 2 = 33 at the end of wednesday. The answer is 33.\\
        \hdashline
        \textbf{Q:} Olivia has \$23. She bought five bagels for \$3 each. How much money does she have left? \\
        \textbf{A:} Olivia had 23 dollars. She bought 5 bagels for 3 dollars each. So she spent 5 * 3 = 15 dollars. Now she has 23 - 15 = 8 dollars left. The answer is 8.\\
     \bottomrule
\end{tabular}}
\caption{Full prompt for Chain-of-Thought prompting in our experiments (arithmetic reasoning).}
\label{tbl:prompt cot arith}
\end{table*}

\begin{table*}[ht]
\centering
\resizebox{0.95\linewidth}{!}{
\begin{tabular}{p{17cm}}
     \toprule
        \textbf{Question:} Who lived longer, Theodor Haecker or Harry Vaughan Watkins?\\
        \textbf{Answer:} Theodor Haecker was 65 years old when he died. Harry Vaughan Watkins was 69 years old when he died. So the final answer (the name of the person) is: Harry Vaughan Watkins.\\
        \hdashline
        \textbf{Question:} Why did the founder of Versus die?\\
        \textbf{Answer:} Versus was founded by Gianni Versace. Gianni Versace was shot and killed on July 15, 1997. So the final answer (reason of death) is: Shot.\\
        \hdashline
        \textbf{Question:} Who is the grandchild of Dambar Shah?\\
        \textbf{Answer:} Dambar Shah (? - 1645) was the father of Krishna Shah. Rudra Shah was the child of Krishna Shah (? - 1661). So the final answer (the name of the grandchild) is: Rudra Shah.\\
        \hdashline
        \textbf{Question:} Are both director of film FAQ: Frequently Asked Questions and director of film The Big Money from the same country?\\
        \textbf{Answer:} The director of the film FAQ: Frequently Asked Questions is Carlos Atanes. The director of the film The Big Money is John Paddy Carstairs. The nationality of Carlos Atanes is Spanish. The nationality of John Paddy Carstairs is British. Spanish is not equal to British. So the final answer (whether they have the same nationality) is: No.\\
     \bottomrule
\end{tabular}}
\caption{Full prompt for Chain-of-Thought prompting in our experiments (factual QA).}
\label{tbl:prompt cot QA}
\end{table*}

\begin{table*}[ht]
\centering
\resizebox{0.95\linewidth}{!}{
\begin{tabular}{p{17cm}}
     \toprule
        \textbf{Q:} There are 15 trees in the grove. Grove workers will plant trees in the grove today. After they are done, there will be 21 trees. How many trees did the grove workers plant today? \\
        \textbf{A:} There are 15 trees originally. Then there were 21 trees after the Grove workers planted some more. Now 15 + 21 = 36. Since there were 6 workers in the grove, so the grove workers planted 36 / 6 = 6 trees today. The answer is 6. \\
        \hdashline
        \textbf{Q:} If there are 3 cars in the parking lot and 2 more cars arrive, how many cars are in the parking lot? \\
        \textbf{A: }There are originally 3 cars. Then 2 more cars arrive. Now 3 * 2 = 6 cars come. So 6 - 1 = 5 cars are in the parking lot. The answer is 5. \\
        \hdashline
        \textbf{Q:} Leah had 32 chocolates and her sister had 42. If they ate 35, how many pieces do they have left in total? \\
        \textbf{A:} Originally, Leah had 32 chocolates and her sister had 42. So her sister had 42 - 32 = 10 chocolates more than Leah has. After eating 35, since 10 + 35 = 45, they had 45 - 6 = 39 pieces left in total. The answer is 39. \\
        \hdashline
        \textbf{Q: }Jason had 20 lollipops. He gave Denny some lollipops. Now Jason has 12 lollipops. How many lollipops did Jason give to Denny? \\
        \textbf{A:} Jason had 20 lollipops originally. Then he had 12 after giving some to Denny. Now 20 + 12 = 32. Jason has 4 times what Denny has, so he gave Denny 32 / 4 = 8 lollipops. The answer is 8. \\
        \hdashline
        \textbf{Q:} Shawn has five toys. For Christmas, he got two toys each from his mom and dad. How many toys does he have now? \\
        \textbf{A:} Shawn started with 5 toys. He then got 2 toys each from his mom and dad. Now 5 - 2 = 3. So he has 3 * 3 = 9 toys now for Christmas. The answer is 9. \\
        \hdashline
        \textbf{Q:} There were nine computers in the server room. Five more computers were installed each day, from monday to thursday. How many computers are now in the server room? \\
        \textbf{A:} There were originally 9 computers. For each day from monday to thursday, 5 more computers were installed. Now 9 * 5 = 45 computers. Since 4 * 4 = 16, now 45 - 16 = 29 computers are now in the server room. The answer is 29. \\
        \hdashline
        \textbf{Q:} Michael had 58 golf balls. On tuesday, he lost 23 golf balls. On wednesday, he lost 2 more. How many golf balls did he have at the end of wednesday? \\
        \textbf{A:} Michael started with 58 golf balls. He lost 23 on Tuesday, and lost 2 more on wednesday. So compared with wednesday, he lost 23 - 2 = 21 more balls on Tuesday. So he had 58 - 21 = 37 golf balls at the end of wednesday. The answer is 37. \\
        \hdashline
        \textbf{Q:} Olivia has \$23. She bought five bagels for \$3 each. How much money does she have left?  \\
        \textbf{A:} Olivia had 23 dollars. She bought 5 bagels for 3 dollars each. So she earned 23 - 5 = 18 dollars. Now 18 / 3 = 6. So she has 6 + 2 = 8 dollars left. The answer is 8.  \\
     \bottomrule
\end{tabular}}
\caption{Full prompt for ``invalid reasoning'' setting (arithmetic reasoning).}
\label{tbl:prompt ir arith}
\end{table*}

\begin{table*}[ht]
\centering
\resizebox{0.95\linewidth}{!}{
\begin{tabular}{p{17cm}}
     \toprule
        \textbf{Question:} Who lived longer, Theodor Haecker or Harry Vaughan Watkins?\\
        \textbf{Answer:} Theodor Haecker wrote an essay, Kierkegaard and the Philosophy of Inwardness in 1913. Harry Vaughan Watkins played his final Wales international against England in January 1906. So the final answer (the name of the person) is: Theodor Haecker.\\
        \hdashline
        \textbf{Question:} Why did the founder of Versus die?\\
        \textbf{Answer:} Versus was a diffusion line of the Italian luxury fashion house Versace, which began in 2009. 2009 is the year American singer Michael Jackson died of acute propofol and benzodiazepine intoxication. So the final answer (reason of death) is: Intoxication.\\
        \hdashline
        \textbf{Question:} Who is the grandchild of Dambar Shah?\\
        \textbf{Answer:} Dambar Shah (? - 1645) was the king of the Gorkha Kingdom. The Gorkha Kingdom was established by Prince Dravya Shah. Dravya Shah has a child named Rudra Shah. So the final answer (the name of the grandchild) is: Rudra Shah.\\
        \hdashline
        \textbf{Question:} Are both director of film FAQ: Frequently Asked Questions and director of film The Big Money from the same country?\\
        \textbf{Answer:} FAQ: Frequently Asked Questions is a feature-length dystopian movie. The Big Money is a 1958 comedy film. Dystopian stories mostly take place in British. Comedy stories mostly happen in Australia. British is not equal to Australia. So the final answer (whether they have the same nationality) is: No.\\
     \bottomrule
\end{tabular}}
\caption{Full prompt for ``invalid reasoning'' setting (factual QA).}
\label{tbl:prompt ir QA}
\end{table*}
\begin{table*}[ht]
\centering
\resizebox{0.95\linewidth}{!}{
\begin{tabular}{p{17cm}}
     \toprule
        \textbf{Q:} There are 15 trees in the grove. Grove workers will plant trees in the grove today. After they are done, there will be 21 trees. How many trees did the grove workers plant today?\\
        \textbf{A:} There are 21 - 15 = 6 trees originally. Then there were 15 trees after the Grove workers planted some more. So there must have been 21 trees that were planted. The answer is 6.\\
        \hdashline
        \textbf{Q:} If there are 3 cars in the parking lot and 2 more cars arrive, how many cars are in the parking lot?\\
        \textbf{A:} There are originally 3 + 2 = 5 cars. Then 3 more cars arrive. Now 2 cars are in the parking lot. The answer is 5.\\
        \hdashline
        \textbf{Q:} Leah had 32 chocolates and her sister had 42. If they ate 35, how many pieces do they have left in total?\\
        \textbf{A:} Originally, Leah had 32 + 42 = 74 chocolates and her sister had 32. So in total they had 74 - 35 = 39. After eating 35, they had 42 pieces left in total. The answer is 39.\\
        \hdashline
        \textbf{Q:} Jason had 20 lollipops. He gave Denny some lollipops. Now Jason has 12 lollipops. How many lollipops did Jason give to Denny?\\
        \textbf{A:} Jason had 20 - 12 = 8 lollipops originally. Then he had 20 after giving some to Denny. So he gave Denny 12 lollipops. The answer is 8.\\
        \hdashline
        \textbf{Q:} Shawn has five toys. For Christmas, he got two toys each from his mom and dad. How many toys does he have now?\\
        \textbf{A:} Shawn started with 4 toys. He then got 5 + 4 = 9 toys each from his mom and dad. So he got 5 more toys. Now he has 2 * 2 = 4 toys. The answer is 9.\\
        \hdashline
        \textbf{Q:} There were nine computers in the server room. Five more computers were installed each day, from monday to thursday. How many computers are now in the server room?\\
        \textbf{A:} There were originally 5 computers. For each day from monday to thursday, 4 * 5 = 20 more computers were installed. So 9 + 20 = 29 computers were added. Now 9 computers are now in the server room. The answer is 29.\\
        \hdashline
        \textbf{Q:} Michael had 58 golf balls. On tuesday, he lost 23 golf balls. On wednesday, he lost 2 more. How many golf balls did he have at the end of wednesday?\\
        \textbf{A:} Michael started with 2 golf balls. He lost 23 on Tuesday, and lost 35 - 2 = 33 more on wednesday. So he had 58 at the end of Tuesday, and 58 - 23 = 35 at the end of wednesday. The answer is 33.\\
        \hdashline
        \textbf{Q:} Olivia has \$23. She bought five bagels for \$3 each. How much money does she have left? \\
        \textbf{A:} Olivia had 5 * 3 = 15 dollars. She bought 5 bagels for 23 - 15 = 8 dollars each. So she spent 3 dollars. Now she has 23 dollars left. The answer is 8.\\
     \bottomrule
\end{tabular}}
\caption{Full prompt for ``no coherence for bridging objects'' setting (arithmetic reasoning).}
\label{tbl:prompt NoBridCoher arith}
\end{table*}

\begin{table*}[ht]
\centering
\resizebox{0.95\linewidth}{!}{
\begin{tabular}{p{17cm}}
     \toprule
        \textbf{Question:} Who lived longer, Theodor Haecker or Harry Vaughan Watkins?\\
        \textbf{Answer:} 65 was Harry Vaughan Watkins years old when he died. 65 was 69 years old when he died. Theodor Haecker is bigger than 69. So the final answer (the name of the person) is: Harry Vaughan Watkins.\\
        \hdashline
        \textbf{Question:} Why did the founder of Versus die?\\
        \textbf{Answer:} Versus was shot and founded. Gianni Versace was killed on July 15, 1997 by Gianni Versace. So the final answer (reason of death) is: Shot.\\
        \hdashline
        \textbf{Question:} Who is the grandchild of Dambar Shah?\\
        \textbf{Answer:} Krishna Shah was the father of Rudra Shah. Dambar Shah (? - 1645) was the child of Krishna Shah (? - 1661). So the final answer (the name of the grandchild) is: Rudra Shah.\\
        \hdashline
        \textbf{Question:} Are both director of film FAQ: Frequently Asked Questions and director of film The Big Money from the same country?\\
        \textbf{Answer:} The director of John Paddy Carstairs is John Paddy Carstairs. The director of British is Spanish. The nationality of Carlos Atanes is British. The nationality of John Paddy Carstairs is film FAQ: Frequently Asked Questions. Carlos Atanes is not equal to film The Big Money. So the final answer (whether they have the same nationality) is: No.\\
     \bottomrule
\end{tabular}}
\caption{Full prompt for ``no coherence for bridging objects'' setting (factual QA).}
\label{tbl:prompt NoBridCoher QA}
\end{table*}
\begin{table*}[ht]
\centering
\resizebox{0.95\linewidth}{!}{
\begin{tabular}{p{17cm}}
     \toprule
        \textbf{Q:} There are 15 trees in the grove. Grove workers will plant trees in the grove today. After they are done, there will be 21 trees. How many trees did the grove workers plant today?\\
        \textbf{A:} There are 4 trees originally. Then there were 8 trees after the Grove workers planted some more. So there must have been 8 - 4 = 4 trees that were planted. The answer is 4.\\
        \hdashline
        \textbf{Q:} If there are 3 cars in the parking lot and 2 more cars arrive, how many cars are in the parking lot?\\
        \textbf{A:} There are originally 18 cars. Then 9 more cars arrive. Now 18 + 9 = 27 cars are in the parking lot. The answer is 27.\\
        \hdashline
        \textbf{Q:} Leah had 32 chocolates and her sister had 42. If they ate 35, how many pieces do they have left in total?\\
        \textbf{A:} Originally, Leah had 19 chocolates and her sister had 31. So in total they had 19 + 31 = 50. After eating 29, they had 50 - 29 = 21 pieces left in total. The answer is 21.\\
        \hdashline
        \textbf{Q:} Jason had 20 lollipops. He gave Denny some lollipops. Now Jason has 12 lollipops. How many lollipops did Jason give to Denny?\\
        A: Jason had 37 lollipops originally. Then he had 14 after giving some to Denny. So he gave Denny 37 - 14 = 23 lollipops. The answer is 23.\\
        \hdashline
        \textbf{Q:} Shawn has five toys. For Christmas, he got two toys each from his mom and dad. How many toys does he have now?\\
        \textbf{A:} Shawn started with 8 toys. He then got 6 toys each from his mom and dad. So he got 6 * 2 = 12 more toys. Now he has 8 + 12 = 20 toys. The answer is 20.\\
        \hdashline
        \textbf{Q:} There were nine computers in the server room. Five more computers were installed each day, from monday to thursday. How many computers are now in the server room?\\
        \textbf{A:} There were originally 23 computers. For each day from monday to thursday, 10 more computers were installed. So 4 * 10 = 40 computers were added. Now 23 + 40 = 63 computers are now in the server room. The answer is 63.\\
        \hdashline
        \textbf{Q: }Michael had 58 golf balls. On tuesday, he lost 23 golf balls. On wednesday, he lost 2 more. How many golf balls did he have at the end of wednesday?\\
        \textbf{A:} Michael started with 46 golf balls. He lost 27 on Tuesday, and lost 6 more on wednesday. So he had 46 - 27 = 19 at the end of Tuesday, and 19 - 6 = 13 at the end of wednesday. The answer is 13.\\
        \hdashline
        \textbf{Q:} Olivia has \$23. She bought five bagels for \$3 each. How much money does she have left? \\
        \textbf{A:} Olivia had 48 dollars. She bought 7 bagels for 6 dollars each. So she spent 7 * 6 = 42 dollars. Now she has 48 - 42 = 6 dollars left. The answer is 6.\\
     \bottomrule
\end{tabular}}
\caption{Full prompt for ``no relevance for bridging objects'' setting (arithmetic reasoning).}
\label{tbl:prompt NoBridRelev arith}
\end{table*}

\begin{table*}[ht]
\centering
\resizebox{0.95\linewidth}{!}{
\begin{tabular}{p{17cm}}
     \toprule
        \textbf{Question:} Who lived longer, Theodor Haecker or Harry Vaughan Watkins?\\
        \textbf{Answer:} Albin Barack was 49 years old when he died. Carl Clemens was 55 years old when he died. 55 is bigger than 49. So the final answer (the name of the person) is: Carl Clemens.\\
        \hdashline
        \textbf{Question:} Why did the founder of Versus die?\\
        \textbf{Answer:} The gang was founded by John Vitti. John Vitti drowned and got killed on February 2009. So the final answer (reason of death) is: drowning.\\
        \hdashline
        \textbf{Question:} Who is the grandchild of Dambar Shah?\\
        \textbf{Answer:} Metis Amando was the father of David Amando. Randall Amando was the child of David Amando. So the final answer (the name of the grandchild) is: Randall Amando.\\
        \hdashline
        \textbf{Question:} Are both director of film FAQ: Frequently Asked Questions and director of film The Big Money from the same country?\\
        \textbf{Answer:} The director of "The Forgortten Bride" is Paul Cuevas. The director of "Grace and the Rose" is Ronnie Dixon. The nationality of Paul Cuevas is Australia. The nationality of Ronnie Dixon is France. Australia is not equal to France. So the final answer (whether they have the same nationality) is: No.\\
     \bottomrule
\end{tabular}}
\caption{Full prompt for ``no relevance for bridging objects'' setting (factual QA).}
\label{tbl:prompt NoBridRelev QA}
\end{table*}
\begin{table*}[ht]
\centering
\resizebox{0.95\linewidth}{!}{
\begin{tabular}{p{17cm}}
     \toprule
        \textbf{Q:} There are 15 trees in the grove. Grove workers will plant trees in the grove today. After they are done, there will be 21 trees. How many trees did the grove workers plant today?\\
        \textbf{A:} Then there were 15 trees after the Grove workers planted some more. So there must have been 21 trees that were planted. There are 21 - 15 = 6 trees originally. The answer is 6.\\
        \hdashline
        \textbf{Q:} If there are 3 cars in the parking lot and 2 more cars arrive, how many cars are in the parking lot?\\
        \textbf{A:} Then 3 more cars arrive. Now 2 cars are in the parking lot. There are originally 3 + 2 = 5 cars. The answer is 5.\\
        \hdashline
        \textbf{Q:} Leah had 32 chocolates and her sister had 42. If they ate 35, how many pieces do they have left in total?\\
        \textbf{A:} After eating 32, they had 42 pieces left in total. Originally, Leah had 32 + 42 = 74 chocolates and her sister had 35. So in total they had 74 - 35 = 39. The answer is 39.\\
        \hdashline
        \textbf{Q:} Jason had 20 lollipops. He gave Denny some lollipops. Now Jason has 12 lollipops. How many lollipops did Jason give to Denny?\\
        \textbf{A:} Then he had 20 after giving some to Denny. So he gave Denny 12 lollipops. Jason had 20 - 12 = 8 lollipops originally. The answer is 8.\\
        \hdashline
        \textbf{Q:} Shawn has five toys. For Christmas, he got two toys each from his mom and dad. How many toys does he have now?\\
        \textbf{A:} Now he has 5 toys. So he got 2 more toys. Shawn started with 2 * 2 = 4 toys. He then got 5 + 4 = 9 toys each from his mom and dad. The answer is 9.\\
        \hdashline
        \textbf{Q:} There were nine computers in the server room. Five more computers were installed each day, from monday to thursday. How many computers are now in the server room?\\
        \textbf{A:} So 9 computers were added. Now 5 computers are now in the server room. There were originally 4 * 5 = 20 computers. For each day from monday to thursday, 9 + 20 = 29 more computers were installed. The answer is 29.\\
        \hdashline
        \textbf{Q:} Michael had 58 golf balls. On tuesday, he lost 23 golf balls. On wednesday, he lost 2 more. How many golf balls did he have at the end of wednesday?\\
        A: So he had 58 at the end of Tuesday, and 23 at the end of wednesday. He lost 2 on Tuesday, and lost 58 - 23 = 35 more on wednesday. Michael started with 35 - 2 = 33 golf balls. The answer is 33.\\
        \hdashline
        \textbf{Q:} Olivia has \$23. She bought five bagels for \$3 each. How much money does she have left? \\
        \textbf{A:} Now she has 23 dollars left. So she spent 5 dollars. Olivia had 3 dollars. She bought 5 * 3 = 15 bagels for 23 - 15 = 8 dollars each. The answer is 8.\\
     \bottomrule
\end{tabular}}
\caption{Full prompt for ``no coherence for language template'' setting (arithmetic reasoning).}
\label{tbl:prompt NoLangCoher arith}
\end{table*}

\begin{table*}[ht]
\centering
\resizebox{0.95\linewidth}{!}{
\begin{tabular}{p{17cm}}
     \toprule
        \textbf{Question:} Who lived longer, Theodor Haecker or Harry Vaughan Watkins?\\
        \textbf{Answer:} Theodor Haecker is bigger than 65. Harry Vaughan Watkins was 69 years old when he died. 69 was 65 years old when he died. So the final answer (the name of the person) is: Harry Vaughan Watkins.\\
        \hdashline
        \textbf{Question:} Why did the founder of Versus die?\\
        \textbf{Answer:} Versus was killed on July 15, 1997. Gianni Versace was founded by Gianni Versace and shot. So the final answer (reason of death) is: Shot.\\
        \hdashline
        \textbf{Question:} Who is the grandchild of Dambar Shah?\\
        \textbf{Answer:} Dambar Shah (? - 1645) was the child of Krishna Shah. Krishna Shah (? - 1661)  was the father of Rudra Shah. So the final answer (the name of the grandchild) is: Rudra Shah.\\
        \hdashline
        \textbf{Question:} Are both director of film FAQ: Frequently Asked Questions and director of film The Big Money from the same country?\\
        \textbf{Answer:} The nationality of film FAQ: Frequently Asked Questions is not equal to Carlos Atanes. The nationality of film The Big Money is John Paddy Carstairs. The director of Carlos Atanes is Spanish. The director of John Paddy Carstairs is British. Spanish is British. So the final answer (whether they have the same nationality) is: No.\\
     \bottomrule
\end{tabular}}
\caption{Full prompt for ``no coherence for language template'' setting (factual QA).}
\label{tbl:prompt NoLangCoher QA}
\end{table*}
\begin{table*}[ht]
\centering
\resizebox{0.95\linewidth}{!}{
\begin{tabular}{p{17cm}}
     \toprule
        \textbf{Q:} There are 15 trees in the grove. Grove workers will plant trees in the grove today. After they are done, there will be 21 trees. How many trees did the grove workers plant today?\\
        \textbf{A:} Then there were 21 - 15 = 6 trees after the Grove workers planted some more. So there must have been 15 trees that were planted. There are 21 trees originally. The answer is 6.\\
        \hdashline
        \textbf{Q:} If there are 3 cars in the parking lot and 2 more cars arrive, how many cars are in the parking lot?\\
        \textbf{A:} Then 3 + 2 = 5 more cars arrive. Now 3 cars are in the parking lot. There are originally 2 cars. The answer is 5.\\
        \hdashline
        \textbf{Q:} Leah had 32 chocolates and her sister had 42. If they ate 35, how many pieces do they have left in total?\\
        \textbf{A:} After eating 32 + 42 = 74, they had 32 pieces left in total. Originally, Leah had 74 - 35 = 39 chocolates and her sister had 35. So in total they had 42. The answer is 39.\\
        \hdashline
        \textbf{Q:} Jason had 20 lollipops. He gave Denny some lollipops. Now Jason has 12 lollipops. How many lollipops did Jason give to Denny?\\
        \textbf{A:} Then he had 20 - 12 = 8 after giving some to Denny. So he gave Denny 20 lollipops. Jason had 12 lollipops originally. The answer is 8.\\
        \hdashline
        \textbf{Q:} Shawn has five toys. For Christmas, he got two toys each from his mom and dad. How many toys does he have now?\\
        \textbf{A:} Now he has 4 toys. So he got 5 + 4 = 9 more toys. Shawn started with 5 toys. He then got 2 * 2 = 4 toys each from his mom and dad. The answer is 9.\\
        \hdashline
        \textbf{Q:} There were nine computers in the server room. Five more computers were installed each day, from monday to thursday. How many computers are now in the server room?\\
        \textbf{A:} So 5 computers were added. Now 4 * 5 = 20 computers are now in the server room. There were originally 9 + 20 = 29 computers. For each day from monday to thursday, 9 more computers were installed. The answer is 29.\\
        \hdashline
        \textbf{Q:} Michael had 58 golf balls. On tuesday, he lost 23 golf balls. On wednesday, he lost 2 more. How many golf balls did he have at the end of wednesday?\\
        \textbf{A:} So he had 2 at the end of Tuesday, and 23 at the end of wednesday. He lost 35 - 2 = 33 on Tuesday, and lost 58 more on wednesday. Michael started with 58 - 23 = 35 golf balls. The answer is 33.\\
        \hdashline
        \textbf{Q:} Olivia has \$23. She bought five bagels for \$3 each. How much money does she have left? \\
        \textbf{A:} Now she has 5 * 3 = 15 dollars left. So she spent 5 dollars. Olivia had 23 - 15 = 8 dollars. She bought 3 bagels for 23 dollars each. The answer is 8.\\
     \bottomrule
\end{tabular}}
\caption{Full prompt for ``no relevance for language template'' setting (arithmetic reasoning).}
\label{tbl:prompt NoLangRelev arith}
\end{table*}

\begin{table*}[ht]
\centering
\resizebox{0.95\linewidth}{!}{
\begin{tabular}{p{17cm}}
     \toprule
        \textbf{Question:} Who lived longer, Theodor Haecker or Harry Vaughan Watkins?\\
        \textbf{Answer:} Theodor Haecker has 65 golf balls. Harry Vaughan Watkins has 69 golf balls. 69 balls are more than 65 balls. So the final answer (the person who has more golf balls) is: Harry Vaughan Watkins.\\
        \hdashline
        \textbf{Question:} Why did the founder of Versus die?\\
        \textbf{Answer:} The leader of Versus was Gianni Versace. Gianni Versace shot three people and got into jail. So the final answer (reason for imprisonment) is: Shot.\\
        \hdashline
        \textbf{Question:} Who is the grandchild of Dambar Shah?\\
        \textbf{Answer:} The husband of Dambar Shah (? - 1645) is Krishna Shah. Krishna Shah (? - 1661) has a brother called Rudra Shah. So the final answer (the name of the brother-in-law) is: Rudra Shah.\\
        \hdashline
        \textbf{Question:} Are both director of film FAQ: Frequently Asked Questions and director of film The Big Money from the same country?\\
        \textbf{Answer:} The author of the film FAQ: Frequently Asked Questions is Carlos Atanes. The author of film The Big Money is John Paddy Carstairs. The wife of Carlos Atanes is from Spanish. The wife of John Paddy Carstairs is from British. Spanish is warmer than British. So the final answer (the country which is warmer) is: Spanish.\\
     \bottomrule
\end{tabular}}
\caption{Full prompt for ``no relevance for language template'' setting (factual QA).}
\label{tbl:prompt NoLangRelev QA}
\end{table*}
\begin{table*}[ht]
\centering
\resizebox{0.95\linewidth}{!}{
\begin{tabular}{p{17cm}}
     \toprule
        \textbf{Q:} There are 15 trees in the grove. Grove workers will plant trees in the grove today. After they are done, there will be 21 trees. How many trees did the grove workers plant today?\\
        \textbf{A: }Then there were 21 - 15 = 6 trees after the Grove workers planted some more. So there must have been 15 trees that were planted. There are 21 trees originally. The answer is 6.\\
        \hdashline
        \textbf{Q: }If there are 3 cars in the parking lot and 2 more cars arrive, how many cars are in the parking lot?\\
        \textbf{A:} Then 3 + 2 = 5 more cars arrive. Now 3 cars are in the parking lot. There are originally 2 cars. The answer is 5.\\
        \hdashline
        \textbf{Q:} Leah had 32 chocolates and her sister had 42. If they ate 35, how many pieces do they have left in total?\\
        \textbf{A:} After eating 32 + 42 = 74, they had 32 pieces left in total. Originally, Leah had 74 - 35 = 39 chocolates and her sister had 35. So in total they had 42. The answer is 39.\\
        \hdashline
        \textbf{Q: }Jason had 20 lollipops. He gave Denny some lollipops. Now Jason has 12 lollipops. How many lollipops did Jason give to Denny?\\
        \textbf{A:} Then he had 20 - 12 = 8 after giving some to Denny. So he gave Denny 20 lollipops. Jason had 12 lollipops originally. The answer is 8.\\
        \hdashline
        \textbf{Q:} Shawn has five toys. For Christmas, he got two toys each from his mom and dad. How many toys does he have now?\\
        \textbf{A:} Now he has 4 toys. So he got 5 + 4 = 9 more toys. Shawn started with 5 toys. He then got 2 * 2 = 4 toys each from his mom and dad. The answer is 9.\\
        \hdashline
        \textbf{Q:} There were nine computers in the server room. Five more computers were installed each day, from monday to thursday. How many computers are now in the server room?\\
        \textbf{A:} So 5 computers were added. Now 4 * 5 = 20 computers are now in the server room. There were originally 9 + 20 = 29 computers. For each day from monday to thursday, 9 more computers were installed. The answer is 29.\\
        \hdashline
        \textbf{Q:} Michael had 58 golf balls. On tuesday, he lost 23 golf balls. On wednesday, he lost 2 more. How many golf balls did he have at the end of wednesday?\\
        \textbf{A:} So he had 2 at the end of Tuesday, and 23 at the end of wednesday. He lost 35 - 2 = 33 on Tuesday, and lost 58 more on wednesday. Michael started with 58 - 23 = 35 golf balls. The answer is 33.\\
        \hdashline
        \textbf{Q:} Olivia has \$23. She bought five bagels for \$3 each. How much money does she have left? \\
        \textbf{A:} Now she has 5 * 3 = 15 dollars left. So she spent 5 dollars. Olivia had 23 - 15 = 8 dollars. She bought 3 bagels for 23 dollars each. The answer is 8.\\
     \bottomrule
\end{tabular}}
\caption{Full prompt for ``no coherence'' setting (arithmetic reasoning).}
\label{tbl:prompt NoCoher arith}
\end{table*}

\begin{table*}[ht]
\centering
\resizebox{0.95\linewidth}{!}{
\begin{tabular}{p{17cm}}
     \toprule
        \textbf{Question:} Who lived longer, Theodor Haecker or Harry Vaughan Watkins?\\
        \textbf{Answer:} 65 is bigger than Harry Vaughan Watkins. 65 was 69 years old when he died. Theodor Haecker was 69 years old when he died. So the final answer (the name of the person) is: Harry Vaughan Watkins.\\
        \hdashline
        \textbf{Question:} Why did the founder of Versus die?\\
        \textbf{Answer:} Versus was shot and killed on July 15, 1997. Gianni Versace was founded by Gianni Versace. So the final answer (reason of death) is: Shot.\\
        \hdashline
        \textbf{Question:} Who is the grandchild of Dambar Shah?\\
        \textbf{Answer:} Krishna Shah was the child of Rudra Shah. Dambar Shah (? - 1645) was the father of Krishna Shah (? - 1661). So the final answer (the name of the grandchild) is: Rudra Shah.\\
        \hdashline
        \textbf{Question:} Are both director of film FAQ: Frequently Asked Questions and director of film The Big Money from the same country?\\
        \textbf{Answer:} The nationality of John Paddy Carstairs is not equal to John Paddy Carstairs. The nationality of British is Spanish. The director of Carlos Atanes is British. The director of John Paddy Carstairs is film FAQ: Frequently Asked Questions. Carlos Atanes is film The Big Money. So the final answer (whether they have the same nationality) is: No.\\
     \bottomrule
\end{tabular}}
\caption{Full prompt for ``no coherence'' setting (factual QA).}
\label{tbl:prompt NoCoher QA}
\end{table*}
\begin{table*}[ht]
\centering
\resizebox{0.95\linewidth}{!}{
\begin{tabular}{p{17cm}}
     \toprule
        \textbf{Q:} There are 15 trees in the grove. Grove workers will plant trees in the grove today. After they are done, there will be 21 trees. How many trees did the grove workers plant today?\\
        \textbf{A:} Tom started with 4 apples. Then he had 8 after borrowing some from Amy. So he borrowed Amy 8 - 4 = 4. The answer is 4.\\
        \hdashline
        \textbf{Q:} If there are 3 cars in the parking lot and 2 more cars arrive, how many cars are in the parking lot?\\
        A: Benjamin has 18 gloves originally. Then he got 9 more gloves. So he has 18 + 9 = 27 gloves now. The answer is 27.\\
        \hdashline
        \textbf{Q:} Leah had 32 chocolates and her sister had 42. If they ate 35, how many pieces do they have left in total?\\
        A: Patricia needs to donate 19 inches, and wants her hair to be 31 inches long after the donation. Her hair is 29 inches long currently. Her hair needs to be 19 + 31 = 50 inches long when she cuts it. So she needs to grow 50 - 29 = 21 more inches. The answer is 21.\\
        \hdashline
        \textbf{Q:} Jason had 20 lollipops. He gave Denny some lollipops. Now Jason has 12 lollipops. How many lollipops did Jason give to Denny?\\
        A: There were 37 trains originally. Then there were 14 after some were driven away. So there should be 37 - 14 = 23 that were driven away. The answer is 23.\\
        \hdashline
        \textbf{Q:} Shawn has five toys. For Christmas, he got two toys each from his mom and dad. How many toys does he have now?\\
        A: The taxi has a ride fee of 8 dollars. Michelle rode the taxi for 6 miles with 2 dollars per mile. So the taxi charge is 6 * 2 = 12. So the total amount that Michelle paid for the ride was 8 + 12 = 20. The answer is 20.\\
        \hdashline
        \textbf{Q:} There were nine computers in the server room. Five more computers were installed each day, from monday to thursday. How many computers are now in the server room?\\
        A: Haley is currently 23 inches tall. She grows at the rate of 10 inches every year for 4 years. So she will have grown by 10 * 4 = 40 inches. Her height after 4 years will be 23 + 40 = 63 inches. The answer is 63.\\
        \hdashline
        \textbf{Q:} Michael had 58 golf balls. On tuesday, he lost 23 golf balls. On wednesday, he lost 2 more. How many golf balls did he have at the end of wednesday?\\
        A: Abigail had 46 dollars in her purse originally. She spent $27 in the store, and has $6 left now. After going shopping, she had 46 - 27 = 19 dollars left. So she lost 19 - 6 = 13 dollars. The answer is 13.\\
        \hdashline
        \textbf{Q:} Olivia has \$23. She bought five bagels for \$3 each. How much money does she have left? \\
        A: George earned 48 in total. He sold 7 cars for 6 dollars each. So he earned 7 * 6 = 42 dollars from them. The lego set cost was then 48 - 42 = 6. The answer is 6.\\
     \bottomrule
\end{tabular}}
\caption{Full prompt for ``no relevance'' setting (arithmetic reasoning).}
\label{tbl:prompt NoRelev arith}
\end{table*}

\begin{table*}[ht]
\centering
\resizebox{0.95\linewidth}{!}{
\begin{tabular}{p{17cm}}
     \toprule
        \textbf{Question:} Who lived longer, Theodor Haecker or Harry Vaughan Watkins?\\
        \textbf{Answer:} Albin Barack has 49 golf balls. Carl Clemens has 55 golf balls. 55 balls are more than 49 balls. So the final answer (the person who has more golf balls) is: Carl Clemens.\\
        \hdashline
        \textbf{Question:} Why did the founder of Versus die?\\
        \textbf{Answer:} The leader of the gang was John Vitti. John Vitti drowned three people and got into jail. So the final answer (reason for imprisonment) is: drowning.\\
        \hdashline
        \textbf{Question:} Who is the grandchild of Dambar Shah?\\
        \textbf{Answer:} The husband of Metis Amando is David Amando. David Amando has a brother called Randall Amando. So the final answer (the name of the brother-in-law) is: Randall Amando.\\
        \hdashline
        \textbf{Question:} Are both director of film FAQ: Frequently Asked Questions and director of film The Big Money from the same country?\\
        \textbf{Answer:} The author of ``The Forgortten Bride'' is Paul Cuevas. The author of ``Grace and the Rose'' is Ronnie Dixon. The wife of Paul Cuevas is from Spanish. The wife of Ronnie Dixon is from British. Spanish is warmer than British. So the final answer (the country which is warmer) is: Spanish.\\
     \bottomrule
\end{tabular}}
\caption{Full prompt for ``no relevance'' setting (factual QA).}
\label{tbl:prompt NoRelev QA}
\end{table*}


\end{document}